# An Evaluation of Support Vector Machines as a Pattern Recognition Tool

**Eugene A. Borovikov**

University of Maryland at College Park

3/13/1999

**Abstract:** The purpose of this report is in examining the generalization performance of Support Vector Machines (SVM) as a tool for pattern recognition and object classification. The work is motivated by the growing popularity of the method that is claimed to guarantee a good generalization performance for the task in hand. The method is implemented in MATLAB. SVMs based on various kernels are tested for classifying data from various domains.

**Keywords:** Statistical Learning Theory, Pattern Recognition, Support Vector Machine, SVM, VC Dimension

## Introduction

Support Vector Machines (SVM) are binary classifiers of objects represented by points in $\mathbf{R}^n$. The foundations of SVM have been developed by Vapnik (1995). These pattern classifiers are gaining popularity due to a number of attractive features including a promising generalization performance. The formulation embodies the Structural Risk Minimization (SRM) principle, as opposed to the Empirical Risk Minimization (ERM) approach commonly employed within statistical learning methods. It is this difference which equips SVMs with an excellent potential to generalize. Since many real-world objects can be represented as points in $\mathbf{R}^n$, and multi-class classifiers can be built by employing arrays of SVMs, the technique is expected to be applicable to a great variety of classification problems. Feasibility and computational cost of such applications are separate issues, and ought to be considered on the case-by-case basis.

## SVM Background

The SVM theory grew out of considerations of under what circumstances, and how quickly, the mean of some empirical quantity converges uniformly, as the number of data points increases, to the true mean. To describe the idea of a Support Vector Machine, we first have to address the issue of Structural Risk Minimization principle. Let us start by posing a generic statistical learning problem.

### *Statistical Learning*

Suppose that we have a feature space $X$ of potential observations; each observation ought to be labeled as one of $K$ classes, or 'doubt' $D$, or 'outlier' $O$; let $Y = \{1,\ldots,K,D,O\}$ be the space of possible decisions, then a *classifier* is a map from $X$ to $Y$. Let the *training set* $T$ be a set of $n$ classified cases; *statistical learning* is then finding a classifying map from $X$ to $Y$ based on data in $T$; any future case (outside $T$) will be *classified* by the map found.

For the purpose of reformulating of a generic problem for an SVM, we let $X = \mathbf{R}^n$ and $Y = \{-1,1\}$. Assume there is an unknown probability distribution $P(\mathbf{x},y)$ from which the labeled observations are drawn. For any mapping $f: X \times A \to Y$, where $A$ is a parameter space, define the following quantity:

$$R(\alpha) = \int \frac{1}{2} |y - f(x,\alpha)| dP(x,y)$$

This quantity has several names. The most common are: expected risk, actual risk, or just the risk. We say that for a fixed parametric family $\{f(\mathbf{x},\alpha)\}$, any choice of a particular $\alpha$ produces a machine (i.e. a classifier). It is our goal to find a machine with the minimal *actual risk*, but since $P(\mathbf{x},y)$ is usually unknown, it is impossible to do directly. It is, however, possible to find an upper bound for the actual risk and pose a problem for its minimization. Prior to doing that, we need to introduce the notion of a function set *capacity* and define some means of measuring that capacity.

The term *capacity* can be introduced as the ability of a machine to learn any training set without an error. Suppose we have a data set of $l$ points that can be assigned labels 1 or -1. Clearly, there are $2^l$ ways to label the data set. If for each labeling, there is a function in the function set $\{f(\mathbf{x},\alpha)\}$ that can correctly assign those labels, we say that the data set is *shattered* by the function set. Maximum cardinality of the data set that can be shattered by $\{f(\mathbf{x},\alpha)\}$ is called the *Vapnik-Chervonenkis (VC) dimension* of that function set. VC-dimension is clearly a property of the function family and can then be used as a measure of capacity of a particular learning machine belonging to that family.

## *Structural Risk Minimization*

Define the *empirical risk* as the measured mean error rate on the training set:

$$R_{emp}(\alpha) = \frac{1}{2l}\sum_{i=1}^{l}|y_i - f(x_i,\alpha)|$$

Notice that no probability distribution appears here; this quantity is fixed for particular $\alpha$ and particular training set $\{\mathbf{x}_i, y_i\}$. Now, we can write the upper bound on the actual risk established by Vapnik:

$$R(\alpha) \le R_{emp}(\alpha) + \sqrt{\frac{h(\log(2l/h)+1) - \log(\eta/4)}{l}}$$

Here $\eta = \Pr\{y = 1\}$ and $h$ is the VC dimension of $\{f(\mathbf{x},\alpha)\}$. The second term in the right-hand side is called *VC confidence*. Finding a learning machine with the minimum upper bound on the actual risk leads us to a method of choosing an optimal machine for a given task. This is the essential idea of the *structural risk minimization* (SRM).

Suppose we have a sequence of nested function families $H_1 \subset H_2 \subset H_3 \subset \ldots$ such that their VC dimensions satisfy $h_1 < h_2 < h_3 < \ldots$ The objective of SRM then is in finding such $H_i$ for which the upper bound on the actual risk is minimal. This can be achieved through the following two-stage process:
- for each $H_i$, identify a machine with minimal $R_{emp}(\alpha)$
- for the final classifier, choose the machine, for which $R_{emp}(\alpha)$+VC-confidence is minimal

## *Support Vector Machines*

One kind of classifiers is especially convenient for implementing the Structural Risk Minimization principle. This kind of machines is called Support Vector Machines (SVM). The origin of the name will be clear later.

### Linear SVM

Given training data as a sequence of $l$ labeled points in $R^n$, the task is to find an "optimal" hyper-plane separating them. There is a factor here that we have to take into account: linear separability of training data.

For the linearly separable case, the support vector algorithm simply looks for the separating hyper-plane with the largest margin, where margin = $2d$ and $d$ is the distance from the hyper-plane to the nearest positive or negative example. It turns out that the margin is inversely proportional to the absolute value of hyper-plane's normal $|\mathbf{w}|$. This reduces the original problem to the following optimization problem:

*minimize* $|\mathbf{w}|$ subject to $y_i(\mathbf{w}\cdot\mathbf{x}_i + b) - 1 \ge 0$,

where $(\mathbf{x}_i, y_i)$ are the training examples and $b$ is the bias term.

A typical solution for the 2D case is shown in Figure 1. The solid line is the solution hyper-plane, the margin is the distance between the two parallel dashed lines, the circled negative and positive examples are called *support vectors*. Notice that the number of the support vectors is usually small comparing to the size of the training set. It is also should be quite obvious that the solution hyper-plane is defined only by the support vectors, that is if we remove all other training points, the hyper-plane will not be affected.

The linearly non-separable case is handled via introduction of slack variables $\xi_i$ to penalize training errors. Then the problem can be rewritten as follows:

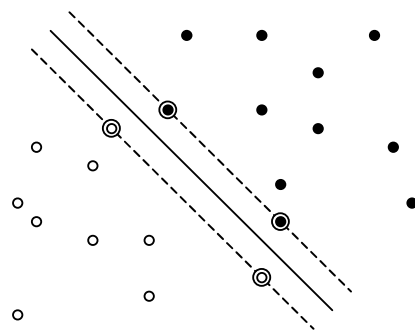

**Figure 1**  An SVM solution to a 2D linearly separable problem

*minimize* $|\mathbf{w}|^2/2 + C(\Sigma_i \xi_i)^k$ subject to:

$$w \cdot x_i + b \geq 1 - \xi_i, \text{ for } y_i = 1$$
$$w \cdot x_i + b \leq -1 + \xi_i, \text{ for } y_i = -1$$
$$\xi_i \geq 0$$

where $C$ is a parameter to be chosen by the user. Large values of $C$ correspond to stronger penalties for errors. Notice that for an error to occur, the corresponding $\xi_i$ must exceed unity, so $\Sigma_i \xi_i$ is an upper bound on the number of training errors.

To solve the above optimization problem, one rewrites it in terms of the positive Lagrange multipliers, $\alpha_i$, one for each constraint. This results in the following Lagrangian:

$$L_P \equiv \frac{1}{2}\|w\|^2 - \sum_{i=1}^{l} \alpha_i y_i (w \cdot x_i + b) + \sum_{i=1}^{l} \alpha_i$$

We must now minimize $L_P$ with respect to $\mathbf{w}$, $b$, and simultaneously require that the derivatives of $L_P$ with respect to all $\alpha_i$ vanish, all subject to $0 \leq \alpha_i \leq C$.

Since the above is a convex problem, it is possible (and more convenient) to solve its dual problem involving a different Lagrangian:

$$L_D \equiv \sum_{i=1}^{l} \alpha_i - \frac{1}{2} \sum_{i,j=1}^{l} \alpha_i \alpha_j y_i y_j (x_i \cdot x_j)$$

Problem now is to

*maximize $L_D$*, subject to $0 \leq \alpha_i \leq C$, $\Sigma_i \alpha_i y_i = 0$.

The solution is then given by

$\mathbf{w} = \Sigma_i \alpha_i y_i \mathbf{x}_i$.

The free coefficient $b$ can be found from

$\alpha_i(y_i(\mathbf{w} \cdot \mathbf{x}_i + b) - 1) = 0$,

for any $i$ such that $\alpha_i$ is not zero.

### Non-linear SVM

The treatment of the most interesting, non-linear, case is based on the idea of injecting the data points into a higher-dimensional Hilbert space via some non-linear mapping, and using the linear support vector algorithm there to separate the training examples. This makes use of the fact that the data appears in the training problem only in the form of dot products.

Suppose there was a map $\Phi: R^d \rightarrow \mathbf{H}$, from our 'data space' into a higher-dimensional Hilbert space $\mathbf{H}$. If there were a function evaluating dot products in $\mathbf{H}$ inexpensively, $K(\mathbf{x}_i, \mathbf{x}_j) = \Phi(\mathbf{x}_i) \cdot \Phi(\mathbf{x}_j)$, then we could train our machine in $\mathbf{H}$ and would never need an explicit form of $\Phi$.

In our algorithm for linear SVM, we substitute all dot products $\mathbf{x}_i \cdot \mathbf{x}_j$ with $K(\mathbf{x}_i, \mathbf{x}_j)$, the *kernel* function, and we will end up with a machine that does a linear separation but in a different Hilbert space. Our classifier then will be concerned with the sign of $f(\mathbf{x})$, where

$$f(x) = \sum_{i=1}^{N_s} \alpha_i y_i \Phi(s_i) \cdot \Phi(x) + b = \sum_{i=1}^{N_s} \alpha_i y_i K(s_i, x) + b$$

$\mathbf{s}_i$ are the support vectors, $N_S$ is the number of support vectors. Note: the existence of a kernel function and an appropriate Hilbert space is problem-dependent and has to be established for each new problem using the theorems that Burges sites in his tutorial.

Here are some kernel functions commonly used for pattern recognition problems:

| | |
|---|---|
| $K(x,y) = (x \cdot y + 1)^p$ | results in a classifier that is a polynomial of degree $p$ |
| $K(x,y) = e^{-\frac{\|x-y\|^2}{2\sigma^2}}$ | gives a Gaussian Radial Basis Function classifier |
| $K(x,y) = \tanh(\kappa x \cdot y - \delta)$ | results in a two-layer sigmoidal neural network |

### SRM by SVM

In the framework of the Structural Risk Minimization (SRM) one is given a series of nested function families whose VC dimensions satisfy SRM requirements. The support vector algorithm chooses an optimal classifier from each family of classifiers given the training set. The next step would then be to identify the SVM with the lowest upper bound for the actual risk.

This process can be illustrated using the SV algorithm with the polynomial kernel function. Let $H_i$ be the space of all polynomials of degree not greater than $i$. It should be quite obvious that $H_i \subset H_k$ and $h_i < h_k$, for $i < k$. For each $H_i$, apply the support vector algorithm with the polynomial kernel function letting $p = i$. This will train an SVM for each $i$ from 1 up to some $n$. Out of these $n$ machines, identify one that minimizes the upper bound on the actual risk. Call this SVM the optimal for the task at hand (according to Burges).

## Applications

In this section, we review results of two SVM applications. One deals with the face detection in black-and-white images. The other came along as a test for comparing two classifiers, and it deals with recognition of handwritten digits. We also discuss a possible symbiosis of SVM with decision trees.

### Face Detection

Edgar Osuna et al [3] have shown how to use a support vector machine for detecting vertically oriented and unoccluded frontal views of human faces in grey level images. They claim that their system handles faces over a wide range of scales and works under different lighting conditions, even with moderately strong shadows.

They train a SVM using a database of face and non-face pixel patterns. The training phase uses an iterative decomposition algorithm. The system detects faces by exhaustively scanning an image for face-like patterns at many possible scales, by dividing the original image into overlapping sub-images and classifying them using the trained SVM to determine the appropriate class (face/non-face). Multiple scales are handled by examining windows taken from scaled versions of the original image.

The system has been tested on two sets of images: set A containing 313 high-quality images with one face per image, and set B containing 23 images of mixed quality with a total of 155 faces. Set A involved 4,669,960 pattern windows, while set B 5,383,682. The reported results were as follows:

| test set | A | B |
|---|---|---|
| detect rate | 97.1% | 74.2% |
| false alarms | 4 | 20 |

### Handwritten Digit Recognition

Schölkopf at al [5] compared SVM with Gaussian kernel to Radial Basis Function Classifiers. For one of the tests, they used US Postal Service Database of 9300 handwritten digits (7300 training, 2000 for testing.) An SV algorithm with standard quadratic programming techniques (conjugate gradient descent) has been used. For the 10-class classifiers, the reported results were as follows:

|  | test error rate |
|---|---|
| clustered centers | 6.7% |
| SV centers | 4.9% |
| SVM | 4.2% |

### SVM and Decision Trees

Bennett and Blue have attempted to develop a method for generating logically simple decision trees with multivariate linear or non-linear decisions [2]. The key idea here was to trade the complexity of the tree to the complexity of the decisions made. This kind of compromise is achieved by using a simple SVM for each decision in the tree.

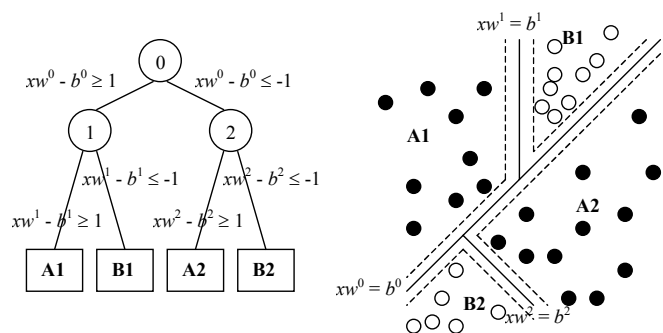

**Figure 2** A symbiosis of SVM and decision tree

A simple example of such a classifier is shown in Figure 2.

# Experiments

In this section we present results of experimental runs of SVM classifiers first on a synthetic dataset, and then on a real-world datasets.

## *Synthetic Problems*

This section describes a number of SVMs that classify data in 2D. All data is synthetic and serves the purpose of demonstrating the SV algorithm's ability to find an optimal hyper surface for the given training data.

The experiment is run as follows. Three different SVMs (Linear, Poly and RBF) are being tested on three different training sets (Linearly separable, Linearly non-separable and Hard-to-separate). Linear SVM uses a linear kernel, Poly uses polynomial kernel of degree 3, and RBF is using an RBF kernel with sigma equal to 10.

As Table 1 shows, the linearly separable case is handled easily by all SVMs. The linearly non-separable case is handled well by Poly and RBF; Linear attempts to do the best it can. The hard-to-separate case is taken well by RBF only; Poly attempts but does not quite separate the data; and Linear is completely lost. This suggests that out of three kinds of kernels, RBF kernel tends to produce the most robust classifiers, but it is the most expensive one of the three. Poly is less expensive and performs reasonably well, but it is naturally limited by the degree of the polynomial used. Linear, the simplest and least expensive of all, inherits all limitations of the Poly because it is a particular case of Poly with degree 1.

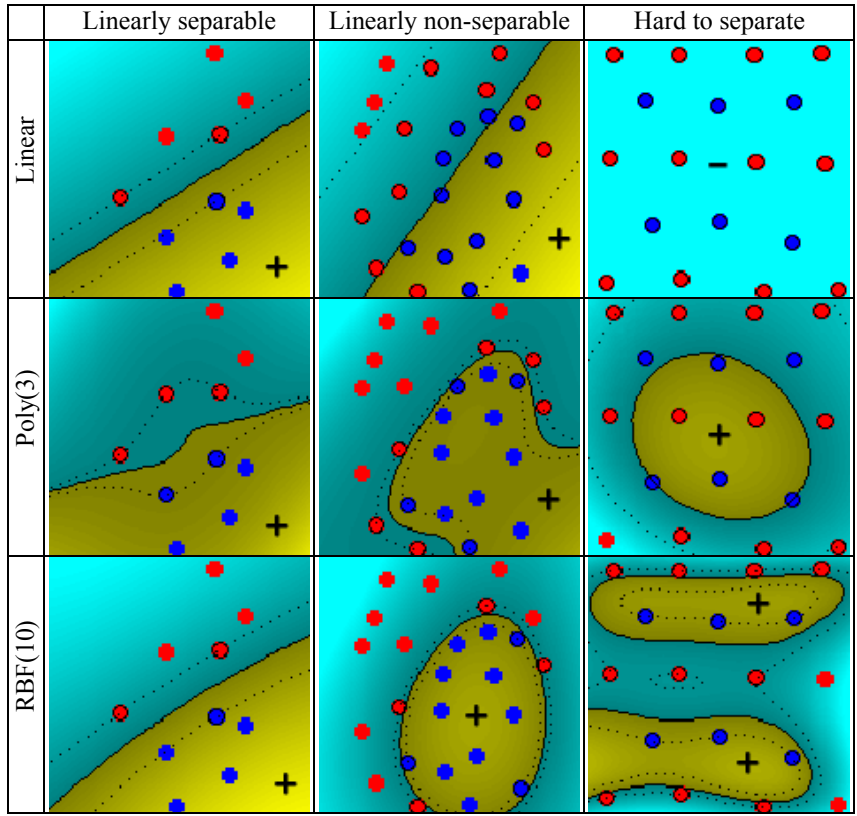

**Table 1:** SVM solutions for synthetic problems

## *Real-world Problems*

In this section, we consider two databases with real-world data. Both databases are taken from the UCI Machine Learning Repository. Both databases contain labeled examples falling in two classes, thus making the application of the SV algorithm straightforward.

### Breast Cancer Classifier

Here we use the Wisconsin Breast Cancer database that contains data on two kinds of cancers (benign and malignant). The class distribution is 65%/35%. Total number of records is 699. Each example has 9 attributes. Missing values are substituted with some numeric values outside the attribute range.

A series of SVM classifiers is trained and tested using different kernel functions. Classifiers are trained on randomly chosen sets of examples of sizes from 100 to 500. Remarkable is the fact, that the straight linear classifier performs quite well having worst-case error rate of 3.9% using a training set of 100 elements (9 support vectors) and best-case 3.14% using a training set of 400 (82 support vectors). Performance of the quadratic-kernel SVM is even better than the linear's. The cubic kernels cause the method to misbehave producing a singularity in the minimization problem. The RBF kernel SVMs have no error no the training data and are more accurate in the best-case (error 1.57% on 500 element training set, 47 support vectors) but are worse than the linear classifier in the worst-case scenario (error 4.9%, 300 element training set, 31 support vectors). On average, however the RBF-kernel SVMs seems to be more robust that the linear SVM. Refer to the "Breast Cancer Classifier runs" in the appendix for the detailed summary.

### Mushroom Classifier

This database presents data on two kinds of mushrooms (edible and poisonous). The class distribution is 52%/48%. Total number of records is 8124. Each example has 22 attributes. Since all attributes are nominal, their values represented by character symbols are taken to be the ASCII codes of those characters.

For this data set, we need to use a robust RBF-kernel SVMs because any polynomial-kernel SVM (including linear) would degenerate in the training phase. As for the Breast Cancer Classifier, SVM classifiers here are trained on randomly chosen sets of examples of sizes from 100 to 500. As the sample run indicates, at first the test error rate is significant (4.7%) and grows for training sets of sizes 100 and 200. But then it sharply drops to 0.8% and continues dropping as the training set size increases. This suggests a superb generalization performance of the SVM. Refer to "Mushroom Classifier runs" in the appendix for more details.

## Conclusions

The objective of this project was to evaluate the Support Vector Machines as a tool for pattern recognition in the framework of statistical learning. The hypothesis under consideration was that Support Vector Machines exhibit an excellent generalization performance, and that they can be successfully applied to a wide range of pattern recognition problems.

We gave an introduction to the Structure Risk Minimization (SRM) principle and provided some background on SV Learning techniques using SRM. We implemented a Support Vector learning system in Matlab and ran a number of experiments using it. The experiments involved both synthetic and real-world data. The results of the experiments revealed to us that Support Vector Machines indeed poses the property of a superb generalization performance and, clearly, can be successfully used in solving various pattern classification tasks.